\documentclass[10pt,twocolumn,letterpaper]{article}
\pdfoutput=1
\usepackage{cvpr}              

\usepackage{graphicx}
\usepackage{amsmath}
\usepackage{amssymb}
\usepackage{booktabs}
\usepackage{multirow}
\usepackage{amsmath}
\usepackage{url}
\usepackage[ruled,linesnumbered]{algorithm2e}

\usepackage[pagebackref,breaklinks,colorlinks]{hyperref}

\usepackage[capitalize]{cleveref}
\crefname{section}{Sec.}{Secs.}
\Crefname{section}{Section}{Sections}
\Crefname{table}{Table}{Tables}
\crefname{table}{Tab.}{Tabs.}

\def\cvprPaperID{2257} 
\def\confName{CVPR}
\def\confYear{2022}

\begin{document}

\title{Open-Vocabulary One-Stage Detection with Hierarchical Visual-Language Knowledge Distillation}

\author{Zongyang Ma\textsuperscript{1,2}, Guan Luo\textsuperscript{1,2}, Jin Gao\textsuperscript{1,2,$\dagger$}, Liang Li\textsuperscript{3,$\dagger$}, Yuxin Chen\textsuperscript{1,2}, Shaoru Wang\textsuperscript{1,2}\\ 
Congxuan Zhang\textsuperscript{4}, and Weiming Hu\textsuperscript{1,2,5}\\
\textsuperscript{1}NLPR, Institute of Automation, Chinese Academy of Sciences\\
\textsuperscript{2}School of Artificial Intelligence, University of Chinese Academy of Sciences\\
\textsuperscript{3}Brain Science Center, Beijing Institute of Basic Medical Sciences ~ \textsuperscript{4}Nanchang Hangkong University\\
\textsuperscript{5}CAS Center for Excellence in Brain Science and Intelligence Technology\\
{\tt\small mazongyang2020@ia.ac.cn, \{gluo,jin.gao\}@nlpr.ia.ac.cn, liang.li.brain@aliyun.com}
}


\maketitle


\begin{abstract}
   Open-vocabulary object detection aims to detect novel object categories beyond the training set. 
   The advanced open-vocabulary two-stage detectors employ instance-level visual-to-visual knowledge distillation to align the visual space of the detector with the semantic space of the Pre-trained Visual-Language Model (PVLM). 
   However, in the more efficient one-stage detector, the absence of class-agnostic object proposals hinders the knowledge distillation on unseen objects, leading to severe performance degradation.
   In this paper, we propose a hierarchical visual-language knowledge distillation method, i.e., HierKD, for open-vocabulary one-stage detection.
   Specifically, a global-level knowledge distillation is explored to transfer the knowledge of unseen categories from the PVLM to the detector. 
   Moreover, we combine the proposed global-level knowledge distillation and the common instance-level knowledge distillation to learn the knowledge of seen and unseen categories simultaneously.
   Extensive experiments on MS-COCO show that our method significantly surpasses the  previous best one-stage detector with 11.9\% and 6.7\% $AP_{50}$ gains under the zero-shot detection and generalized zero-shot detection settings, and reduces the $AP_{50}$ performance gap from 14\% to 7.3\% compared to the best two-stage detector. 
   Code will be released at this url \footnote{\url{ https://github.com/mengqiDyangge/HierKD}}.
%

   
\end{abstract}

\section{Introduction}
\label{sec:intro}
\vspace{-1em}

\newcommand\blfootnote[1]{%
\begingroup 
\renewcommand\thefootnote{}\footnote{#1}%
\addtocounter{footnote}{-1}%
\endgroup 
}
{
	\blfootnote{$^\dagger$ Corresponding authors.
	}
}

\begin{figure}[t]
\centering
\small
\includegraphics[scale=0.29]{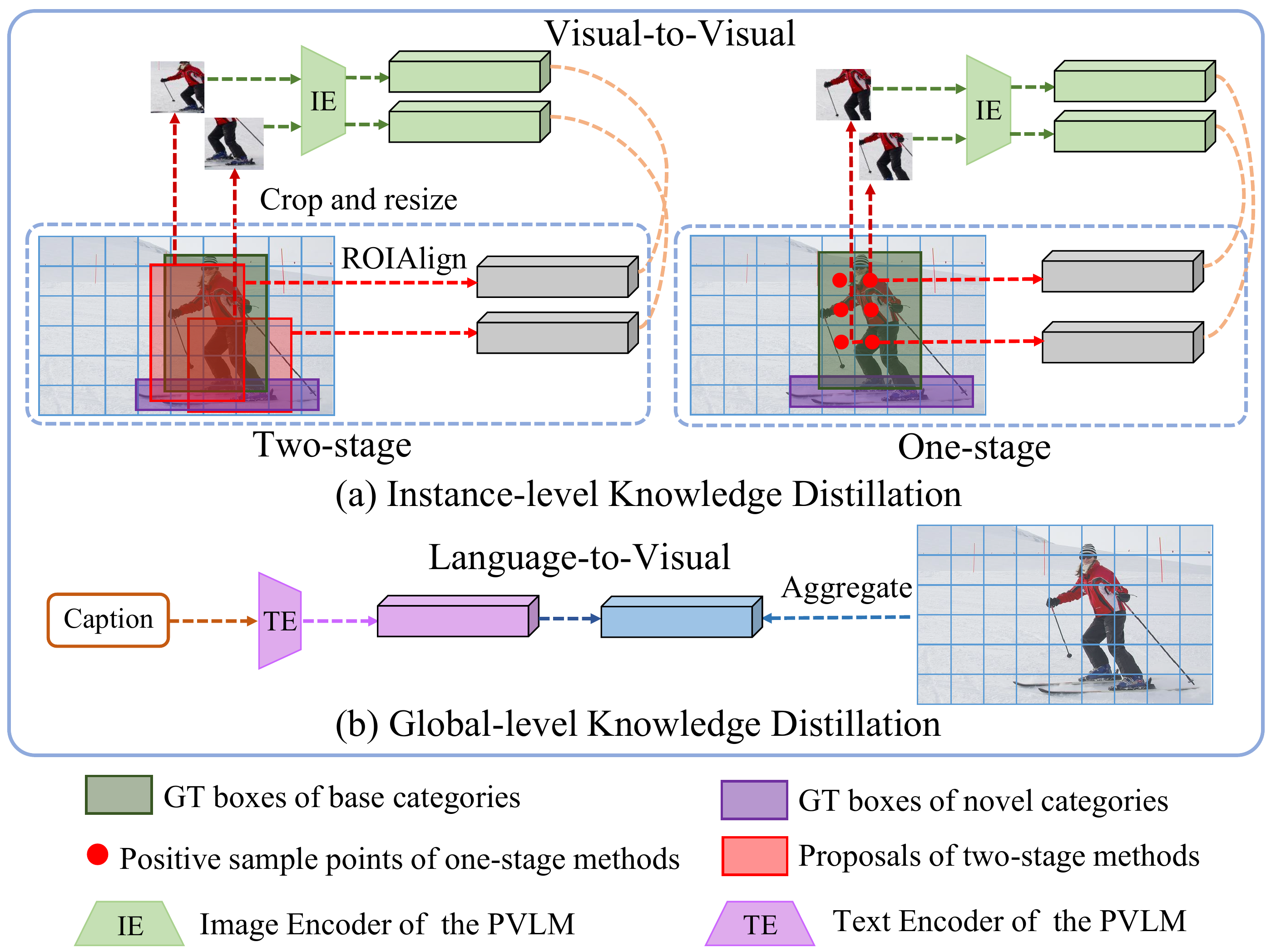}
\caption{\textbf{Comparisons between instance-level and global-level knowledge distillation:} (a) illustrates the pipelines of two-stage methods and one-stage methods with instance-level knowledge distillation.  (b) illustrates our proposed global-level knowledge distillation, which directly distills the caption representation from PVLM to the global image representation from detector.
}
\label{fig:area}
\vspace{-2em}
\end{figure}
%

The emerging trends in advanced detectors \cite{ren2015faster, redmon2016you, redmon2017yolo9000, redmon2018yolov3, bochkovskiy2020yolov4, Lin_2017_CVPR, Lin_2017_ICCV, tian2019fcos, Zhang_2020_CVPR} have improved the speed and accuracy of traditional object detection tasks significantly, whereas the categories they can recognize are limited. Once the traditional detectors are expected to detect more object categories in the real-world scenarios, the usual solution falls on labeling more categories of objects in training sets. However, the cost may be unaffordable and the long-tail distribution will be exacerbated by increasing the unseen categories linearly according to Zipf's law\cite{saichev2009theory}. To overcome these limitations, zero-shot \cite{bansal2018zero} and open-vocabulary \cite{zareian2021open} object detection tasks are proposed to recognize objects from unseen categories (novel categories) while the detector is only trained with annotations from seen categories (base categories). The main difference between these two tasks is that the open-vocabulary detector might have seen a novel object during training though its instance-level annotation is not available. Therefore, the open-vocabulary detector \cite{zareian2021open, xie2021zsd, gu2021zero} has developed more rapidly recently, and their performance also lead the former by a large margin.


There have been some works attempting to redesign the traditional detectors to accomplish the above two detection tasks. These works can also be divided into two-stage \cite{bansal2018zero, li2019zero, zheng2021zero, zheng2020background, gu2021zero, zareian2021open} methods and one-stage \cite{rahman2020improved, zhao2020gtnet, zhu2020don ,xie2021zsd} methods as in traditional detection. It is known that the traditional state-of-the-art one-stage detectors have comparable performance and more concise pipeline compared to traditional two-stage ones. However, in the open-vocabulary object detection, the current best two-stage method ViLD \cite{gu2021zero} significantly surpasses the similar one-stage method \cite{xie2021zsd}. As such, it is encouraging to analyze the reason behind this phenomenon and find ways to narrow this performance gap, and then construct a high-performance open-vocabulary one-stage detector.

We show pipelines of recent two-stage and one-stage open-vocabulary detection methods in Figure \ref{fig:area} (a). It can be seen that both of them perform Instance-level visual-to-visual Knowledge Distillation (IKD) on possible instances of interest in the images. The key difference lies in the selection of instances, i.e., object proposals for two-stage methods and positive sample points for one-stage methods. Compared to the object proposals, there are severe inherent limitations in the positive sample points. We argue that these limitations cause the performance gap between two-stage and one-stage methods.

Specifically, as illustrated in Figure \ref{fig:area} (a), the positive sample points (red points) only cover the area of the objects from base categories (green boxes), so the one-stage methods can only learn the semantic knowledge about the base categories from the PVLM during the distillation. On the contrary, the class-agnostic proposals (red boxes) in two-stage methods usually cover the regions of the objects from novel categories (purple boxes), which enables the two-stage methods to implicitly learn the semantic knowledge of novel categories from the PVLM (See sec \ref{sec:rpn} for a clearer analysis).  
This advantage can effectively expand the semantic category space and further improve performance. What's more, the number of positive sample points is much less than the object proposals in most images, and each positive sample point only covers a smaller area on the feature maps than the proposals. This sparse sampling of the feature map areas during distillation also makes the semantic supervision from PVLM shrink a lot in one-stage methods.

To compensate for these inherent limitations, a straightforward approach is to make use of more sample points of the feature maps for knowledge distillation. Thus, in this work, we propose a weakly supervised global-level language-to-visual knowledge distillation method (GKD) to achieve this approach.  As shown in Figure \ref{fig:area} (b), GKD exploits the visual captions that potentially contain semantic knowledge of novel categories, and performs language-to-visual knowledge distillation between caption representation and global-level image representation. In this way, GKD implicitly aligns all sample points in the image with the caption semantics, so that the sample points belonging to the novel categories can also learn their related semantic knowledge from the PVLM.

Finally, our proposed GKD is combined with the commonly used IKD to perform open-vocabulary one-stage detection in an end-to-end fashion, leading to a hierarchical knowledge distillation mechanism-based detector, namely HierKD. We summarize our contributions as follows: 



 \begin{itemize}
     \item A weakly supervised global-level language-to-visual knowledge distillation method is explored to learn novel category knowledge beyond training labels for one-stage detection. 
     \item An end-to-end hierarchical visual-language knowledge distillation mechanism is proposed to achieve a high-performing open-vocabulary one-stage detector.  
     \item The proposed HierKD detector significantly surpasses the previous best open-vocabulary one-stage detector with 11.9\% and 6.7\% $AP_{50}$ gains under the zero-shot detection and generalized zero-shot detection settings respectively on MS-COCO dataset. 
 \end{itemize}

\section{Related Work}
\label{sec:rela}


\begin{figure*}[t]
\centering
\includegraphics[scale=0.665]{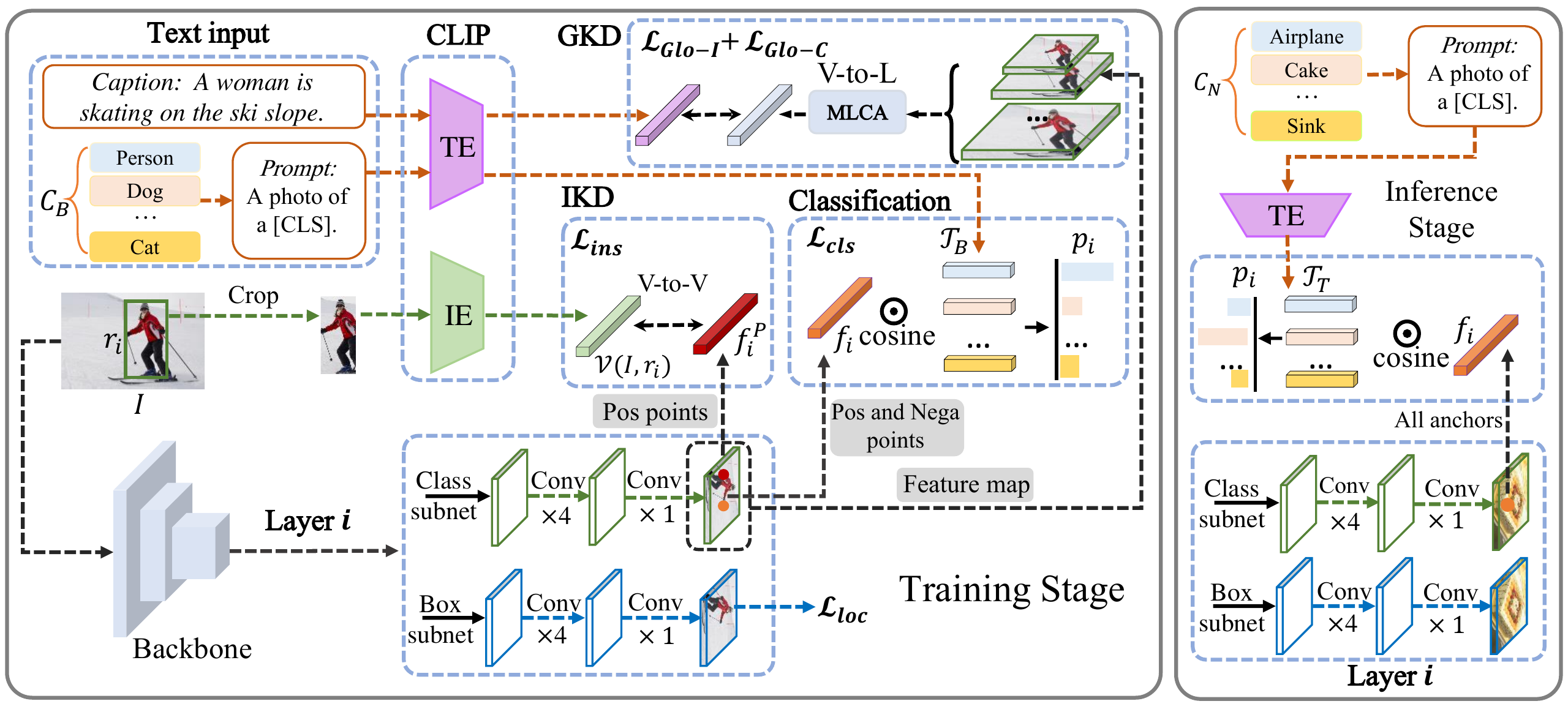}
\caption{\textbf{Overview of our open-vocabulary one-stage detector with hierarchical visual-language knowledge distillation:} In the training stage, the classification branch is initialized with the CLIP textual embedding of base categories. For IKD, the aim is to minimize the distance between the features of sparse positive sample points on feature maps and the CLIP visual embedding of the cropped regions. The GKD aggregates all the multi-layer feature maps to directly align with the captions by cross attention. During inference, the knowledge distillation modules are removed and the CLIP textual embedding is initialized with the novel categories. It is noteworthy that the distillation has less impact on the regression branch thanks to the inherent characteristics of disentanglement in one-stage detectors.}
\label{fig:model}
\vspace{-1.2em}
\end{figure*}

\noindent \textbf{Zero-shot Learning:}  As the capability of image recognition with supervised learning has reached a high-level status, researchers begin to explore how well the classification models can recognize objects of novel categories beyond training sets, which is usually referred as zero-shot learning (ZSL). The earliest works start from modeling the attributes of objects by encoding the label space with binary attribute vectors for recognizing objects \cite{farhadi2009describing, palatucci2009zero, jayaraman2014zero}, while the later works focus more on the semantic representation of the visual space \cite{frome2013devise, norouzi2013zero, wang2018zero}. Recently, PVLM, e.g., CLIP \cite{radford2021learning}, learns to model visual concepts based on the natural language like human and acquires powerful zero-shot recognition ability. Different from these image-level zero-shot recognition works, we aim at exploring the open-vocabulary instance-level detectors. Nevertheless, PVLM is also closely related to our work for we hope to transfer its zero-shot recognition ability to the open-vocabulary detection by knowledge distillation.     

\noindent \textbf{Zero-shot and Open-vocabulary Detection:} Zero-shot and open-vocabulary detection both focus on designing a detector which can recognize and localize objects of novel categories beyond the training sets. Some works explore the two-stage detectors \cite{bansal2018zero,zareian2021open,xie2021zsd, zheng2021zero, li2019zero, zheng2020background} and have achieved the state-of-the-art performance. 
Zareian \etal \cite{zareian2021open} designed a projection layer for aligning visual space with textual semantic space based on PixelBERT \cite{huang2020pixel}. 
Xie \etal \cite{xie2021zsd} proposed to distill region-level visual features from CLIP. 
Another direction focuses on designing more efficient one-stage detectors by modifying loss functions \cite{rahman2020improved}, introducing transductive learning \cite{rahman2019transductive}, and synthesizing features for unseen objects \cite{zhu2020don}. 
Gu \etal \cite{gu2021zero} also distilled knowledge from CLIP with a baseline one-stage detector YOLO-v5 \cite{bochkovskiy2020yolov4}. A
lthough it significantly surpasses the previous one-stage methods, there is still a large performance gap compared to the advanced two-stage methods. We have analyzed the reason behind the poorly performing one-stage methods during instance-level knowledge distillation, and concentrate on compensating for their inherent limitations.

\section{Approach}
\label{sec: method}
Figure \ref{fig:model} illustrates the overall framework of our proposed open-vocabulary one-stage detector HierKD. It consists of a teacher pre-trained visual-language model and a student detector during the training phase. Here we employ a pre-trained visual-language model named CLIP \footnote{CLIP ViT-B/32 is selected for fair comparisons with other methods.} for its superior performance. The student model aims to learn the teacher model's zero-shot recognition ability by our proposed hierarchical visual-language knowledge distillation mechanism. In particular, the positive sample points learn from Image Encoder (IE) of the teacher model by instance-level visual-to-visual knowledge distillation, and the multi-scale feature maps from the detector directly transfer knowledge from Text Encoder (TE) of teacher by global-level language-to-visual knowledge distillation.


\noindent \textbf{Notations:} The categories in the training set, \ie, base categories is denoted as $C_B$, and the novel categories in the testing set is denoted as $C_N$. In addition, TE and IE of CLIP are denoted as $\mathcal{T}$ and $\mathcal{V}$, respectively. The textual embedding $\mathcal{T}_{B}$ used in training is initialized offline by feeding each category in $C_B$ with a prompt, \ie ``a photo of a [CLS].", into the text encoder $\mathcal{T}$. During inference, the only modification is to replace $C_B$ with $C_N$ or the union $C_B \cup C_N$ under different settings.




\subsection{Choosing and Modifying a Base Detector}
\label{sec: base-model}
    The first challenge is how to adapt an off-the-shelf one-stage base detector to the open-vocabulary object detection task with necessary structural modifications. 

\noindent \textbf{Choosing a Base One-stage Detector:} We first leverage ATSS \cite{Zhang_2020_CVPR} as the base one-stage detector for two reasons: (1) The adaptive training sample selection mechanism makes it a top performer in the traditional object detection task; (2) There is only one anchor at each location on the feature maps, which is important because modifying the classification layer (see below) will dramatically increase memory consumption as the number of anchors increases. 

\noindent \textbf{Modifying the Base Detector:} We then make two modifications to the original ATSS, as illustrated in Figure \ref{fig:model}:
(1) The original convolution-based classification layer is modified to the classification form of CLIP with the names or descriptions of the dataset’s categories embedded by TE. A background embedding $\mathcal{T}_{bg}$ is also required since this modification would lose the original detector's ability to distinguish the background samples.  The $\mathcal{T}_{bg}$ \footnote{We also try to randomly initialize it \cite{xie2021zsd} and set a fixed zero vector with a bias \cite{zareian2021open}, but we finally get similar performance.} is initialized by feeding ``a photo of background." into $\mathcal{T}$, which allows to learn the background in the training stage. The sigmoid function is also replaced with the softmax function, and the final classification loss is based on the softmax focal loss.  
\begin{equation}
 \begin{aligned}
    & p_i = \mathrm{SoftMax}([\tau_c \cdot (\mathcal{T}_B f_i^T), \tau_c \cdot (\mathcal{T}_{bg} f_i^T)])~,\\
    & \mathcal{L}_{cls} = \frac{1}{N_{pos}} \sum_{i=0}^{N}\mathcal{L}_{\rm{focal loss}}(p_i, y_i)~,\\
 \end{aligned}
\end{equation}
where $f_i$, $p_i$ and $y_i$ denote the anchor feature, classification result and label of the anchor respectively. $\tau_c$ is a learnable temperature coefficient during training, and $N_{pos}$ is the number of positive sample points while $N$ denotes the total number of positive and negativet samle points;
(2) The centerness branch in ATSS is replaced with an IOU branch \cite{paa-eccv2020} for mitigating the misalignment between classification task and regression task to a certain extent.  


\subsection{Instance-level Knowledge Distillation}
\label{sec: ins-level}
We then introduce the instance-level knowledge distillation, which aims at transferring knowledge from the image encoder $\mathcal{V}$. Following the common practice, only the features of positive samples are fetched for distillation. Since the positive sample points in ATSS may have relatively small IOU values with respect to the ground-truth boxes, we set a fixed IOU threshold to further filter out the positive samples with small IOUs and acquire the features for the remaining positive sample points $\{f_1, f_2, ... , f_{N_{pos}}\}$. Unlike ZSD-YOLO \cite{xie2021zsd}, we use the predicted boxes of regression branch instead of the ground-truth boxes to crop regions from image $I$ for the sake of data augmentation. These cropped regions are then resized to 224 $\times$ 224 to adapt to the input image size of $\mathcal{V}$. We use a resizing method that can keep more image information, \ie, ``Long side + padding", which resizes the long side to 224 and pads the short side with 0. Next, the features to be mimicked $\{ \mathcal{V}(I, r_i), r_i \in \mathcal{R} \}$ can be obtained by feeding these resized regions $\mathcal{R}$ into the image encoder $\mathcal{V}$. Finally, the knowledge is transferred from the CLIP image encoder to detectors with distillation as follows:
\begin{equation}
    \mathcal{L}_{ins} = \frac{1}{N_{pos}} \sum_{i=1}^{N_{pos}} \left\|\frac{f_i}{\|f_i\|}_2 -  \frac{\mathcal{V}(I, r_i)}{\|\mathcal{V}(I, r_i)\|}_2\right\|_1~.
\end{equation}

We have also tried with $L_2$ norm for mimicking, and there are no obvious differences among different measures after adjusting the appropriate loss weights.

\subsection{Global-level Knowledge Distillation}
To overcome the limitation of only learning from the base categories, a weakly supervised GKD module is explored by exploiting the image captions to learn the semantic knowledge of novel categories beyond training labels. GKD mimics the contrastive learning in CLIP to match the image-caption pairs and aims at transferring CLIP's large-scale semantic knowledge to the one-stage detector. 

\begin{figure}[t]
\centering
\includegraphics[scale=0.48]{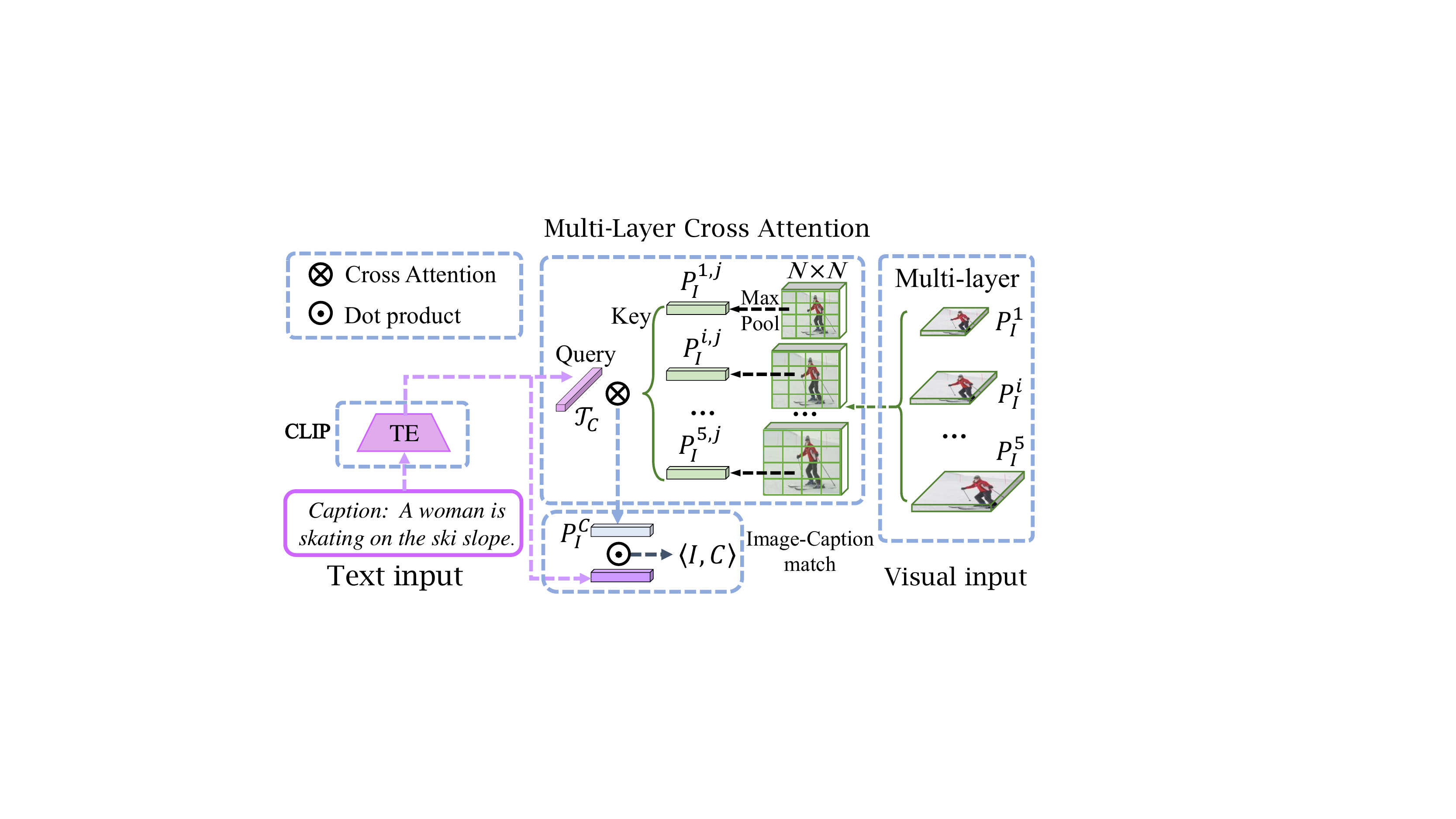}
\caption{\textbf{Global-level knowledge distillation:} This GKD module takes the caption as textual input and the feature maps from multi layers as visual input, and learns to match the image-caption pairs by mimicking the contrastive learning in CLIP through the multi-layer cross-attention.}
\label{fig:global}
\vspace{-1.2em}
\end{figure}

 Figure \ref{fig:global} illustrates the overall process of GKD. Specifically, an arbitrary image denoted by $I$ and its paired caption denoted by $C$ are matched by Multi-Layer Cross Attention (MLCA). For the visual input, feature maps from different FPN layers are evenly divided into $N \times N$ patches, and the Max Pooling operation is performed inside all patches of different feature maps to obtain the patch-level representations. The set of pooled patch features is denote by $\{P^{i,j}_I | i=1,2,3,4,5, j=1,...,N \times N\}$, where $i$ indicates the FPN layer and $j$ is the patch location on the feature maps of each layer. 
 Next, for the textual input, the whole caption $C$ is encoded directly by text encoder $\mathcal{T}$ to represent the textual feature $\mathcal{T}_C$. As the CLIP model is great at extracting the overall high-level textual feature while not at word-level representation in some simple visual-grounding experiments, we choose to take the feature of the entire caption instead of each word.
 
 
After obtaining the textual feature and the set of multi-layer patch features, the cross attention takes these multi-modal inputs to aggregate the patch features. Specifically, the caption is regarded as query, all patches are regarded as keys, and the response between the query and each key can be calculated via cosine similarity. Hence, the aggregation of all patch features is obtained with the normalized similarities as follows:
\begin{equation}
    \begin{aligned}
       e_{i,j} &= \frac{\mathcal{T}_C \cdot P^{i,j}_I}{\| \mathcal{T}_C \|\| P^{i,j}_I \|}~,\\
       P^C_I &= \sum_{i=j=1}^{5, k}\frac{exp(e_{i,j})}{\sum_{i'=j'=1}^{5,k}exp(e_{i',j'})}P^{i,j}_I~,\\
    \end{aligned}
\end{equation}
where $P^C_I$ represents the caption-aware visual feature aggregation, and $e_{i,j}$ is the response between the caption and the $j_{th}$ patch of $i_{th}$ layer. Finally, the matching score $\langle I, C \rangle$ between the image-caption pair $(I, C)$ is:
\begin{equation}
    \langle I, C \rangle = \frac{P^C_I \cdot \mathcal{T}_C}{\| P^C_I \| \| \mathcal{T}_C \|}~. 
\end{equation}

Since the aim of our global-level knowledge distillation is to transfer CLIP's large-scale semantic knowledge to the detector, it is naturally to mimic the contrastive learning in CLIP and also the recent self-supervised learning works \cite{he2020momentum, chen2020simple}. The paired images and captions are regarded as positive pairs in a batch while the others are negative pairs. We introduce a symmetrical contrastive loss function to push the positive pairs and pull the negatives in semantic space:
\begin{equation}
\label{eqa: eq6}
    \begin{aligned}
       \mathcal{L}_{Glo-I} &= -log \frac{exp(\tau_m \cdot \langle I, C \rangle)}{\sum_{C_i = 1}^bexp(\tau_m \cdot \langle I, C_i \rangle)}~, \\
       \mathcal{L}_{Glo-C} &= -log \frac{exp(\tau_m \cdot \langle I, C \rangle)}{\sum_{I_i = 1}^bexp(\tau_m \cdot \langle I_i, C \rangle)}~, \\
    \end{aligned}
\end{equation}
where $\tau_m$ is a trainable temperature coefficient, and $b$ denotes the batch size.

Finally, the hierarchical knowledge distillation of our one-stage detector can be formulated by combining the instance-level knowledge distillation and global-level knowledge distillation:
\begin{equation}
\begin{aligned}
    \mathcal{L} = &\lambda_{cls} \mathcal{L}_{cls} + \lambda_{loc} \mathcal{L}_{loc} + \lambda_{ins} \mathcal{L}_{ins}  \\
     &+ \lambda_{Glo} (\mathcal{L}_{Glo-I} +  \mathcal{L}_{Glo-C})~.
\end{aligned}
\end{equation}

\subsection{Sampling the Negative Samples}
Advanced one-stage detectors often combine focal Loss \cite{Lin_2017_ICCV} or its variants \cite{li2020generalized, li2021generalized, Feng_2021_ICCV} with all negative samples to solve the imbalance problem between positive and negative samples. However, this setting is troubling in open-vocabulary detection, for the detectors will identify more foreground regions as background when generalizing to novel categories in experiments. On the other hand, sampling the negative samples to 1:1 with positive samples as in two-stage methods will boost the performance on novel categories, whereas it seriously affects the base categories. To make a trade-off between the above options, we adopt a sampling strategy by sampling 10\% negative samples to boost the recall performance on novel categories while maintaining the performance on base categories. 

\subsection{Direct Inference Alternative with CLIP}
As the zero-shot recognition ability of the proposed method is transferred from CLIP, we can thus measure the mimicking ability of our method by comparing the performance gap between our model and this CLIP direct inference. We design a simple CLIP direct inference way in algorithm \ref{alg1}. Essentially, it compares the difference in the classification results of the same sample points between the detector and the CLIP.

\vspace{-0.4em}

\begin{algorithm}
\caption{CLIP Direct Inference}
\label{alg1}
\KwIn{CLIP image encoder $\mathcal{V}$ and text encoder $\mathcal{T}$, novel categories $C_N$, trained model $\mathcal{M}$, test images $\mathcal{D}_{T}$}
\KwOut{Detection boxes $B$}
$\mathcal{T}_N \leftarrow \mathcal{T}(Prompt(C_N))$ and normalize\;
\For{$I \in D_T$}
{$A_I \leftarrow \mathcal{M}_{anchor}(I)$\;
$A_I^{fore} \leftarrow \arg\max_{k}(\mathcal{M}_{cls}(A_I) \times \mathcal{M}_{IOU}(A_I))$\; 
$\mathcal{V}_I^{fore} \leftarrow \mathcal{V}(I, \mathcal{M}_{loc}(A_I^{fore})) $ and normalize\;
$S_I^{fore} \leftarrow Softmax(\tau \cdot \mathcal{T}_N \mathcal{V}_I^{fore})$\;
$B \leftarrow B \cup NMS(S_I^{fore}$, $box_I^{fore}$)\;}
\end{algorithm}
\vspace{-2.0em}


\section{Experiments and Results}
\subsection{Dataset and Evaluation Protocol}
We validate our method on the MS-COCO 2017 benchmark under both zero-shot detection (ZSD) and generalized zero-shot detection (GZSD) settings. In the previous ZSD literature, two different types of base/novel split settings are available: the 48/17 and the 65/15 base/novel splits by Bansal et al. \cite{bansal2018zero} and Rahman et al. \cite{rahman2020improved}, respectively. We evaluate both split settings in this paper. Our data preprocessing is the same as Rahman et al. \cite{zareian2021open}. Following the most previous ZSD methods, we evaluate our method using mAP and Recall@100 at IOU=0.5, and mainly focus on the performance of novel categories.  

\subsection{Implementation Details}
Our implementation and hyper-parameter settings are based on MMdetection \cite{chen2019mmdetection}. A standard ResNet-50 \cite{he2016deep} is adopted as the backbone, and all hyper-parameters remain the default settings unless otherwise specified. We set the thresholds of NMS and classification score to 0.4 and 0.0 respectively. The temperature coefficients $\tau_c$ and $\tau_m$ are initialized to 100 and 10 respectively. We also add a gradient clip at 10.0 during the training stage. For the knowledge distillation, the teacher model CLIP is frozen, and the feature maps of different FPN layers are divided into 3 $\times$ 3 patches. We train the model on 4 Tesla V100 GPUs and use a batch size of 16 in IKD and 32 in GKD and HierKD. The learning schedule follows the traditional object detection settings.


\begin{table}[]
\centering
\footnotesize
\setlength{\tabcolsep}{1.8mm}{
\begin{tabular}{c|c|ccc}
\hline
IOU & Base/Novel & AR@100 & AR@300 & AR@1000 \\ \hline \hline
0.5 & 48/17 & 61.9 & 76.9 & 87.5 \\
0.75 & 48/17 & 37.4 & 48.1 & 57.4 \\ \hline
\end{tabular}
}
\caption{Generalization ability of RPN}
\label{tab:rpn_ge}
\vspace{-0.8em}
\end{table}

\begin{table}[]
\footnotesize
\centering
\setlength{\tabcolsep}{1.8mm}{
\begin{tabular}{@{}cccc|cc@{}}
\hline
 Norm & Weight & Region & Area & $AR_{50}$ & $AP_{50}$ \\
 \hline \hline
 $L_1$ & 1 & \textit{pred} & 1$\times$ & 62.4 & 14.6 \\
 $L_2$ & 1 & \textit{pred} & 1$\times$ & \textbf{65.1} & 12.8 \\
 $L_2$ & 10 & \textit{pred} & 1$\times$ & 63.6 & 14.6 \\
 $L_1$ & 1 & \textit{GT} & 1$\times$ & 62.8 & 14.5 \\
 $L_1$ & 1 & \textit{pred} & 1.5$\times$ & 64.5 & \textbf{15.3} \\
 \hline
\end{tabular}%
}
\caption{Comparisons between different sub-module options in IKD. \textit{pred} and \textit{GT} mean cropping regions from prediction boxes and ground-truth boxes, respectively. 1$\times$ and 1.5$\times$ represent cropping the original box and its 1.5$\times$ center expansion respectively.}
\vspace{-1.6em}
\label{tab:ins-abla}
\end{table}

\subsection{Test on generalization ability of RPN}
\label{sec:rpn}
To more clearly illustrate the generalization ability of the RPN, we train the RPN on the base categories and directly transfer it to test on the novel categories.
As shown in Table \ref{tab:rpn_ge}, the category-agnostic proposals in RPN of two-stage methods usually cover the regions of the novel objects, and AR is still up to 37.4 when generating 100 proposals and IOU=0.75, which contributes to feature learning on novel categories during knowledge distillation.

\subsection{Ablation Study}
We conduct ablation studies on the MS-COCO ZSD benchmark to verify the effectiveness of design choices. 
All the results are reported on the novel categories under the 48/17 base/novel split setting unless otherwise specified.

\noindent \textbf{Instance-level Knowledge Distillation:} We compare the impact of different sub-module options in the instance-level knowledge distillation in Table \ref{tab:ins-abla}. Compared to our distillation using $L_1$ norm, replacing it with $L_2$ loss norm will cause a 1.8\% $AP_{50}$ drop, and this gap can be reduced through increasing the loss weight. It is essentially because the distance between the features measured by the L2 norm requires a larger weight to be consistent with the result of the L1 norm. The cropped region factor used in knowledge distillation is not sensitive to using prediction boxes or ground-truth boxes. However, it can improve performance by cropping the 1.5$\times$ expanded box area to provide more contextual information. 

\begin{table}[]
\centering
\footnotesize
\setlength{\tabcolsep}{1.8mm}{
\begin{tabular}{@{}cccc|cc@{}}
\hline
 \textit{Patch} & \textit{Pool} & \textit{Loss} & \textit{bs/gpu} & $AR_{50}$ & $AP_{50}$ \\
 \hline \hline
 \textit{4} & \textit{Ave} & \textit{CL} & \textit{8} & 59.2 & 12 \\
 \textit{4} & \textit{Max} & \textit{CL} & \textit{8} & 64.2 & 20.1 \\
 \textit{3} & \textit{Max} & \textit{CL} & \textit{8} & 61.1 & \textbf{20.7} \\
 \textit{8} & \textit{Max} & \textit{CL} & \textit{8} & 60.8 & 13.7 \\
 \textit{3} & \textit{Max} & \textit{PL} & \textit{8} & 60.9 & 17.9 \\
 \textit{3} & \textit{Max} & \textit{CL} & \textit{4} & \textbf{65.6} & 20.5 \\
 \hline
\end{tabular}%
}
\caption{Comparisons between different sub-module options in GKD. \textit{Ave} and \textit{Max} represent using Average Pooling and Max Pooling to obtain patch features respectively. \textit{CL} denotes training with the contrastive learning loss, while \textit{PL} only considers the cosine similarities between positive pairs. \textit{bs/gpu} is the batch size on each GPU during training.}
\label{tab: glo-abla}
\end{table}


\begin{table}[]
\centering
\footnotesize
\setlength{\tabcolsep}{0.8mm}{
\begin{tabular}{ccc|ccccc}
\hline
IKD & GKD & $\rm IOU_b$ & $AR_{50}$ & $AP_{50}$ & $AP_S$ & $AP_M$ & $AP_L$ \\
\hline \hline
- & - & - & 52.4 & 10.2 & 8.8 & 12.5 & 12.8      \\ \hline
$\surd$ &  &  & 62.4 & 14.6 & 10.1 & 13.2 & 19.1  \\
& $\surd$ &  & 61.1 & 20.7 & 10.1 & 28.5 & 27.5     \\
$\surd$ & $\surd$ &  & 70.1 & 20.7 & 11.5 & 30.2 & 27.0      \\
$\surd$ & $\surd$ & $\surd$ & \textbf{71.3} & \textbf{21.6} & \textbf{11.6} & \textbf{30.7} & \textbf{28.1}   \\
\hline
\end{tabular}%
}
\caption{Verify the effectiveness and compatibility of each module. The first row is the baseline, which is the exploited base detector trained with only classification loss and localization loss.}
\label{tab:module-abla}
\vspace{-1.4em}
\end{table}

\noindent \textbf{Global-level Knowledge Distillation:} As shown in Table \ref{tab: glo-abla}, the different choices of sub-modules have great impacts on the performance. First, We observe that the $AP_{50}$ achieved by using Average Pooling is only about half of Max Pooling. This is caused due to the loss of distinguishability of the patch features obtained through Average Pooling. Moreover, compared to dividing the feature maps into a small number of patches, such as 3$\times$3 or 4$\times$4, dividing it into more patches, such as 8$\times$8, brings a significant $AP_{50}$ drop. It can be attributed to the reason that more training iterations are required to converge for more patches. Additionally, there is no obvious difference between 8 \textit{bs/gpu} and 4 \textit{bs/gpu}. We infer that both of them we can afford are too small for contrastive learning to make a difference. Finally, replacing contrastive learning with only pushing the positive pairs brings a 2.8\% $AP_{50}$ drop. This shows that contrastive learning can better transfer the zero-shot recognition capability of the PVLM.

\noindent \textbf{Distillation Module Analysis:} We quantitatively verify the effectiveness of each distillation module and the compatibility of different modules. We additionally report the detection performance on small objects $AP_S$, medium objects $AP_M$ and large objects $AP_L$ to perform a more detailed analysis. As shown in Table \ref{tab:module-abla}, by adding IKD and GKD to the baseline, we can obtain 4.4\% and 10.5\% $AP_{50}$ gains as well as 10.0\% and 8.7\% $AR_{50}$ gains respectively. This validates the effectiveness of each distillation module. 
In addition, compared to applying IKD and GKD separately, the combination of IKD and GKD, \ie, HierKD, further brings 7.7\% and 9.0\% $AR_{50}$ gains respectively. This shows that the great compatibility of IKD and GKD. The $AP_S$ and $AP_M$ in HierKD are improved by 1.4\% and 1.7\% compared to GKD, which shows that HierKD has advantages in detecting small and medium objects.  
Finally, using IOU branch leads to more improvements on the medium and large objects than the small. It may be because objects with low classification scores and high IOUs generally do not appear on small objects.

\begin{figure}[t]
\centering
\small
\includegraphics[scale=0.40]{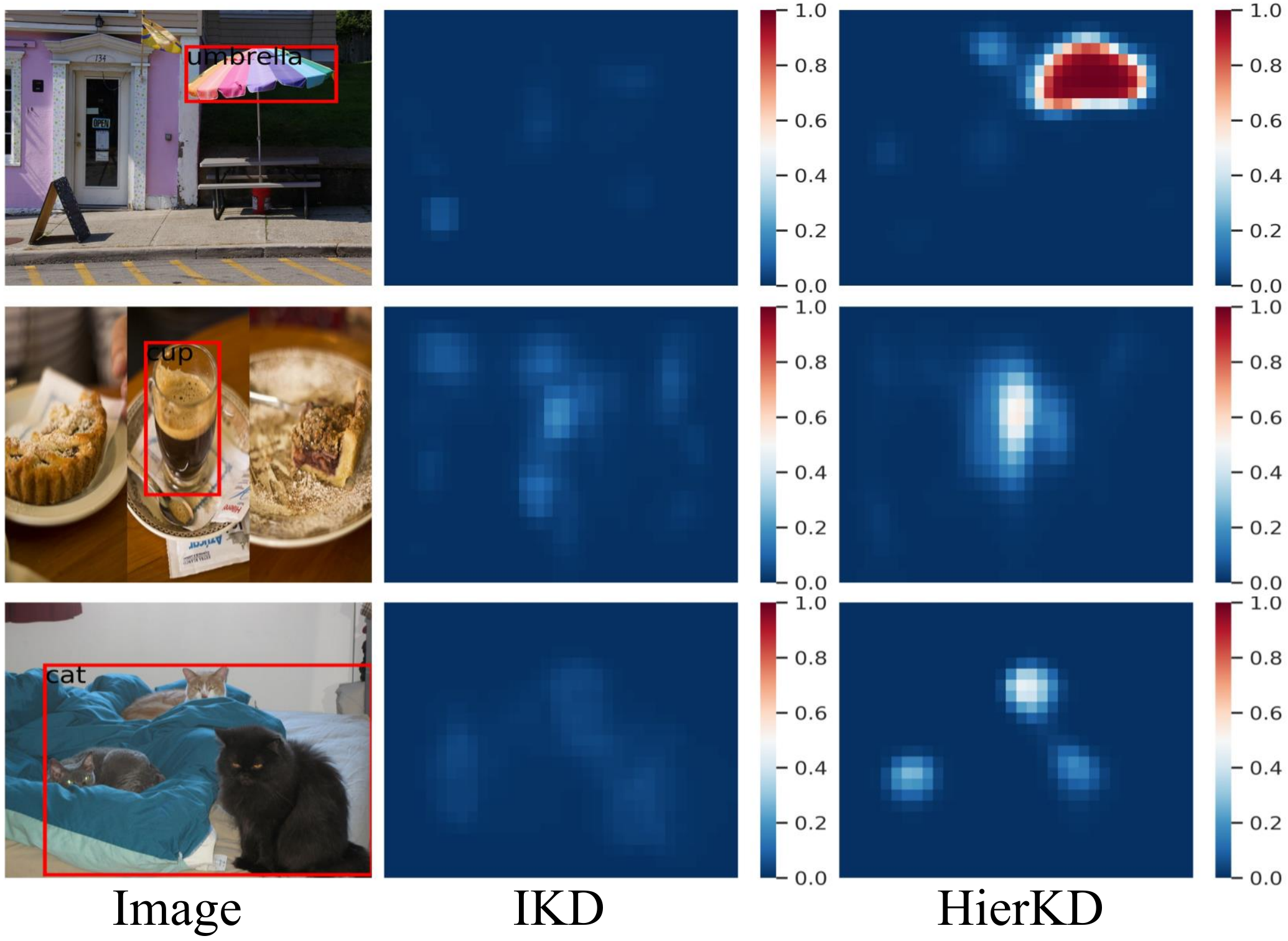}
\caption{Spatial distribution of classification score. The red boxes in the images are ground-truth of the novel categories. The heatmaps in IKD and HierKD show the classification score of anchors at each location for the categories in the red boxes.}
\label{fig:score}
\vspace{-0.4em}
\end{figure}

\begin{table}[]
\footnotesize
\centering
\resizebox{\linewidth}{!}{%
\begin{tabular}{@{}c|cc|cc|cc@{}}
\hline
\multirow{2}{*}{Negative samples} & \multirow{2}{*}{IKD} & \multirow{2}{*}{GKD} & \multicolumn{2}{c|}{Base} & \multicolumn{2}{c}{Novel} \\
     &   &   & $AR_{50}$ & $AP_{50}$ & $AR_{50}$ & $AP_{50}$ \\ 
\hline \hline
1:1   & $\surd$ &  & 71.0 & 37.0 & \textbf{63.0} & \textbf{16.8} \\
10\%  & $\surd$ &  & \textbf{75.9} & 44.3 & 62.4 & 14.6 \\
100\% & $\surd$ &  & 74.5 & \textbf{44.4} & 60.3 & 9.0 \\ 
\hline
1:1   &  & $\surd$ & 69.2 & 34.9 & 60.2 & 19.3 \\
10\%  &  & $\surd$ & \textbf{74.0} & \textbf{42.7} & \textbf{61.1} & \textbf{20.7} \\
100\% &  & $\surd$ & 72.4 & 42.6 & 56.4 & 18.7 \\ 
\hline
\end{tabular}%
}
\caption{Comparisons between different sampling strategies for negative samples. 1:1, 10\%, 100\% mean sampling the same number of negative samples as the positive samples, sampling 10 \% of the negative samples, and using all the negative samples.}
\label{tab: sample-abla}
\vspace{-1.4em}
\end{table}

\begin{figure}[t]
\centering
\includegraphics[scale=0.39]{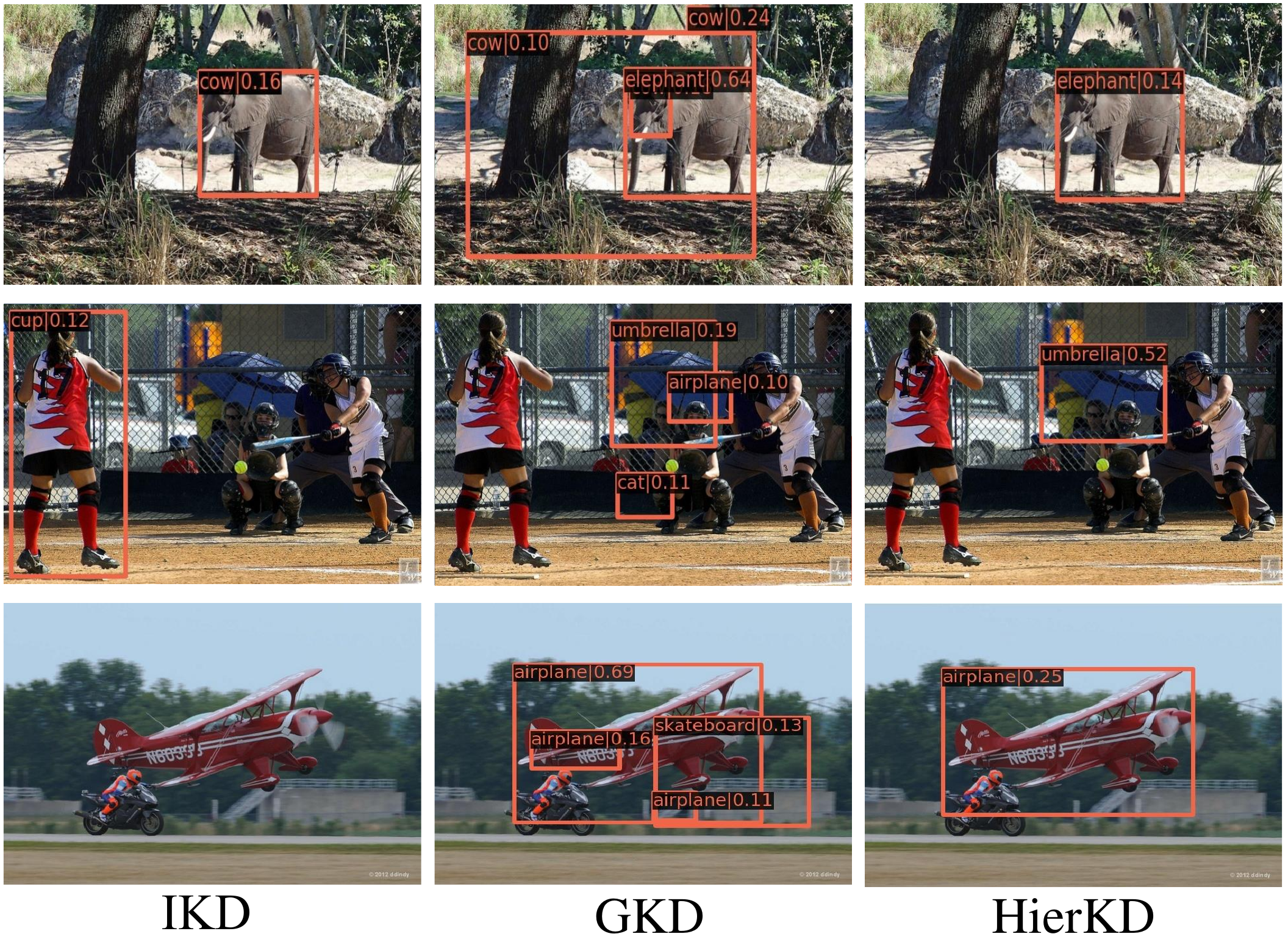}
\caption{Visualization of some detections on novel categories.}
\label{fig:vis}
\end{figure}

We also visualize some classification score distribution and detection results for qualitative analysis. Figure \ref{fig:score} illustrates the spatial distribution of classification score in IKD and HierKD, respectively. We can see that IKD often fails to identify the objects of novel categories, \eg, the ``umbrella'' in the first row. In addition, IKD may also have low confidence in recognition of the objects of novel categories, such as the ``cup'' in the second row and the ``cat'' in the third row. By introducing GKD, the proposed HierKD can recognize the ``umbrella'' in the first row, and also significantly increases the confidence in recognition of the ``cup'' in the second row and the ``cat'' in the third row. This shows that our HierKD can better transfer the novel category knowledge from CLIP and reduce missed detections while increasing detection confidence. 
We also show some detection results of novel categories in Figure \ref{fig:vis}. First, it can be seen that GKD and HierKD can identify more objects of novel categories compared to IKD, such as the ``umbrella'' in the second row and the ``airplane'' in the third row. Moreover, GKD and HierKD also have higher classification accuracy, such as correctly classifying the ``elephant'' in the first row instead of recognizing it as a ``cow'' like IKD. Compared to GKD, HierKD can suppress more meaningless detection results, such as the multiple partial ``airplane'' in the third row. 

\noindent \textbf{Sampling the Negative Samples:} The impact of the sampling strategy for negative samples is shown in Table \ref{tab: sample-abla}. Taking 100\% sampling as the baseline, we can see that 10\% sampling does not cause a large $AP_{50}$ drop on the base categories in comparison with 1:1 sampling. 
When generalizing to novel categories, the $AP_{50}$ obtained by 10\% sampling is not much worse than the 1:1 sampling in IKD while achieving the best in GKD. This validates the effectiveness of the 10\% sampling strategy.

\begin{table}[]
\centering
\footnotesize
\setlength{\tabcolsep}{1.5mm}{
\begin{tabular}{@{}c|cc|ccccc@{}}
\hline
 & Model & CLIP & $AR_{50}$ & $AP_{50}$ & $AP_S$ & $AP_M$ & $AP_L$ \\
\hline \hline
\multirow{2}{*}{IKD} & $\surd$ & &  62.4 & 14.6 & 10.1 & 13.2 & 19.1 \\
  &  & $\surd$ &  \textbf{66.6} & \textbf{24.9} & \textbf{18.3} & \textbf{28.3} & \textbf{32.3}\\ \hline
\multirow{2}{*}{GKD}     & $\surd$ & &  61.1 & 20.7 & 10.1 & \textbf{28.5} & 27.5      \\
  &  & $\surd$  &  \textbf{64.5} & \textbf{23.3} & \textbf{18.2} & 27.9 & \textbf{30.3}\\
 \hline
\multirow{2}{*}{HierKD} & $\surd$ & &  \textbf{70.1} & 20.7 & 11.5 & \textbf{30.2} & 27.0\\
 & & $\surd$  &  65.8 & \textbf{22.8} & \textbf{18.8} & 27.4 & \textbf{29.9}  \\ \hline
\end{tabular}
}
\caption{Comparison with the direct inference with CLIP. Model and CLIP represent inference with the detector from distillation and direct inference with CLIP respectively.}
\label{tab: per-gap}
\vspace{-1.4em}
\end{table}

\begin{table}[]
\footnotesize
\centering
\setlength{\tabcolsep}{0.9mm}{
\begin{tabular}{ccc|c|c|ccc}
\hline
\multicolumn{3}{c|}{\multirow{2}{*}{Method}} & \multirow{2}{*}{Base/Novel} & ZSD & \multicolumn{3}{c}{GZSD} \\ \cline{5-8} 
\multicolumn{3}{c|}{} &  & Novel & Base & Novel & All \\ \hline
\hline
\multicolumn{1}{c|}{\multirow{7}{*}{TS/MS}} & \multicolumn{1}{c|}{\multirow{5}{*}{ZS}} & SB\cite{bansal2018zero} & 48/17 & 0.70 & 29.2 & 0.31 & 24.9 \\
\multicolumn{1}{c|}{} & \multicolumn{1}{c|}{} & LAB\cite{bansal2018zero} & 48/17 & 0.27 & 20.8 & 0.22 & 18.0 \\
\multicolumn{1}{c|}{} & \multicolumn{1}{c|}{} & DESE\cite{bansal2018zero} & 48/17 & 0.54 & 26.7 & 0.27 & 22.1 \\
\multicolumn{1}{c|}{} & \multicolumn{1}{c|}{} & BLC\cite{zheng2020background} & 48/17 & 9.9 & 42.1 & 4.50 & 32.3 \\
\multicolumn{1}{c|}{} & \multicolumn{1}{c|}{} & BA-RPN*\cite{zheng2021zero} & 48/17 & 11.4 & 46.5 & 4.83 & 35.6 \\ \cline{2-8} 
\multicolumn{1}{c|}{} & \multicolumn{1}{c|}{\multirow{2}{*}{OV}} & OVR-CNN\cite{zareian2021open} & 48/17 & 16.7 & - & - & 34.3 \\
\multicolumn{1}{c|}{} & \multicolumn{1}{c|}{} & ViLD*\cite{gu2021zero} & 48/17 & - & 59.5 & 27.6 & 51.3 \\ \hline
\multicolumn{1}{c|}{\multirow{4}{*}{OS}} & \multicolumn{1}{c|}{\multirow{2}{*}{ZS}} & PL*\cite{rahman2020improved} & 48/17 & 10.0 & 35.9 & 4.12 & 27.9 \\
\multicolumn{1}{c|}{} & \multicolumn{1}{c|}{} & DELO\cite{zhu2020don} & 48/17 & 7.6 & 13.8 & 3.41 & 13.0 \\ \cline{2-8} 
\multicolumn{1}{c|}{} & \multicolumn{1}{c|}{\multirow{2}{*}{OV}} & ZSD-YOLO*\cite{xie2021zsd} & 48/17 & 13.4 & 31.7 & 13.6 & 27.0 \\
\multicolumn{1}{c|}{} & \multicolumn{1}{c|}{} & \textbf{HierKD(ours)} & 48/17 & \textbf{25.3} & \textbf{51.3} & \textbf{20.3} & \textbf{43.2} \\ \hline
\multicolumn{1}{c|}{\multirow{2}{*}{TS/MS}} & \multicolumn{1}{c|}{\multirow{2}{*}{ZS}} & BLC\cite{zheng2020background} & 65/15 & 13.1 & 36.0 & 13.1 & 31.7 \\
\multicolumn{1}{c|}{} & \multicolumn{1}{c|}{} & BA-RPN*\cite{zheng2021zero} & 65/15 & 13.6 & 38.7 & 13.6 & 34.0 \\ \hline
\multicolumn{1}{c|}{\multirow{3}{*}{OS}} & \multicolumn{1}{c|}{ZS} & PL*\cite{rahman2020improved} & 65/15 & 12.4 & 34.1 & 12.4 & 30.0 \\ \cline{2-8} 
\multicolumn{1}{c|}{} & \multicolumn{1}{c|}{\multirow{2}{*}{OV}} & ZSD-YOLO*\cite{xie2021zsd} & 65/15 & 18.3 & 31.7 & 17.9 & 29.2 \\
\multicolumn{1}{c|}{} & \multicolumn{1}{c|}{} & \textbf{HierKD(ours)} & 65/15 & \textbf{27.4} & \textbf{48.9} & \textbf{20.4} & \textbf{43.6} \\ \hline
\end{tabular}
}
\caption{\textbf{Comparison with other state-of-the-art methods:} * denotes the state-of-the-art methods in various settings. ``TS/MS'' and ``OS'' are abbreviation of two-stage/multi-stage and one-stage detectors, respectively. ``ZS'' and ``OV'' indicate that the models belong to zero-shot and open-vocabulary detectors, respectively.}
\label{tab:com-sota}
\vspace{-1.8em}
\end{table}

\noindent \textbf{Compared to Direct Inference with CLIP:} The performance gap between the proposed method and direct inference with CLIP is shown in Table \ref{tab: per-gap}. The IKD baseline has only about half of the $AP_{50}$ compared to direct inference with CLIP on all sizes of objects, while our proposed GKD achieves similar performance on medium and large objects. The final HierKD has higher $AR_{50}$ than direct inference with CLIP. However, the $AP_S$ in HierKD lags behind the direct inference with CLIP a lot, which shows that our method has insufficient learning ability for small objects.

\noindent \textbf{Different Training Settings:} As shown in Table \ref{tab:gene_impro}, extending the period of training schedule from 1$\times$ to 2$\times$, 3$\times$, introducing scale jitter (480-800), and changing backbone to larger ResNet-101 can improve the performance of both the base and novel categories. This validates that the proposed HierKD is compatible with the general detection performance improvement techniques.

\subsection{Comparison with the Start-of-the-Art}
We compare our HierKD with the other two-stage methods and one-stage methods on the MS-COCO benchmark in Table \ref{tab:com-sota}, all metrics reported in Table \ref{tab:com-sota} are $AP_{50}$.
\textbf{Limitation:} we can not make a completely fair comparison like the traditional object detection because the factors of batch size, scale jitter, \etc, used in some works (such as ViLD \cite{gu2021zero}) are different from the general settings.

We can observe that under the 48/17 base/novel split setting, HierKD achieves 25.3\% $AP_{50}$ on novel categories under the ZSD setting. HierKD significantly outperforms the previous best one-stage method ZSD-YOLO with 11.9\% $AP_{50}$ gains, and also exceeds the most recent two-stage method OVD (trained without external Conceptual Caption dataset \cite{sharma2018conceptual}) by 8.6\% $AP_{50}$. Under the GZSD setting, HierKD outperforms ZSD-YOLO with 6.7\% gains on novel categories. HierKD also reduces the $AP_{50}$ performance gap from 14\% to 7.3\% compared to the best two-stage method ViLD. Under the GZSD setting, the $AP_{50}$ of HierKD on the novel categories is 5\% lower than that of the ZSD. This is caused by the detection confidence of the novel categories is lower than that of the base categories, so some detection results of novel categories are suppressed during NMS. 

Under another 65/15 base/novel split setting, HierKD surpasses the previous best method ZSD-YOLO with 10.1\% and 2.5\% $AP_{50}$ gains on novel categories under ZSD and GZSD settings respectively. 

\begin{table}[]
\centering
\smaller
\setlength{\tabcolsep}{1.2mm}{
\begin{tabular}{@{}ccc|cc|cc@{}}
\hline
\multirow{2}{*}{Backbone} & \multirow{2}{*}{Schedule} & \multirow{2}{*}{Scale Jitter} & \multicolumn{2}{c|}{Base} & \multicolumn{2}{c}{Novel} \\
          &    &  & $AR_{50}$ & $AP_{50}$ & $AR_{50}$ & $AP_{50}$ \\
\hline \hline
ResNet-50  & 1$\times$ &         & 74.8 & 44.7 & 71.3 & 21.6 \\
ResNet-50  & 2$\times$ &         & 77.5 & 49.0 & 69.8 & 23.1 \\
ResNet-50  & 3$\times$ & $\surd$ & 80.0 & 51.8 & 70.0 & 25.3 \\
ResNet-101 & 3$\times$ & $\surd$ & \textbf{80.8} & \textbf{53.5} & \textbf{71.4} & \textbf{27.3} \\ \hline
\end{tabular}%
}
\caption{Verification of the compatibility with general detection performance improvement techniques.}
\label{tab:gene_impro}
\vspace{-1.0em}
\end{table}

\begin{table}[]
\centering
\footnotesize
\setlength{\tabcolsep}{1.2mm}{
\begin{tabular}{@{}c|c|ccccc@{}}
\hline
            & Base/Novel & $AR_{50}$ & $AP_{50}$ & $AP_S$ & $AP_M$ & $AP_L$ \\
\hline \hline
HierKD      & 48/17      & \textbf{71.4} & 27.3 & 11.4 & 39.5 & 37.3 \\
Upper Bound & 48/17      & 70.7 & \textbf{68.0} & \textbf{36.3} & \textbf{74.5} & \textbf{87.4} \\ \hline
\end{tabular}%
}
\caption{Comparison with the ideal upper bound. All reported metrics are results on the novel categories.}
\label{tab:upper-bound}
\vspace{-1.8em}
\end{table}

\subsection{Upper Bound Analysis}
We can get the ideal upper bound of this type of distillation method by directly using CLIP to classify the instances in the ground-truth boxes and then evaluating the detection results, \ie, the classification results of ground-truth boxes. As shown in Table \ref{tab:upper-bound}, our method achieves a relatively high recall, while the total $AP_{50}$ and $AP$ on objects of various sizes, \ie $AP_S$, $AP_M$, $AP_M$ are still far from the upper bound. This shows that there is still much room to improve the mimicking ability of the proposed HierKD. In addition, this also reminds us of using techniques such as prompt learning \cite{zhou2021coop} to improve the zero-shot recognition ability of CLIP itself, thereby further improving the upper bound of model performance.
\vspace{-0.6em}

\section{Conclusion}
In this work, we have developed a hierarchical visual-language knowledge distillation method,  namely HierKD, to obtain a top-performing one-stage open-vocabulary detector. HierKD uses image caption to distill knowledge in a language-to-visual manner. The rich vocabulary in captions enables HierKD to transfer the semantic knowledge of novel categories from CLIP during training. The results indicate that the proposed HierKD can identify novel objects more accurately and confidently, and significantly surpasses the previous methods. In the future, 
we will continue to explore more efficient and advanced distillation methods to transfer the zero-shot recognition ability of teacher models.
%

\noindent \textbf{Acknowledgment} This work was supported by the National Key R$\&$D Program of China (Grant No. 2018AAA0102803, 2018AAA0102800), the Natural Science Foundation of China (Grant No. U2033210, 62172413, 61972394, 62036011, 62192782, 61721004), the Key Research Program of Frontier Sciences, CAS (Grant No. QYZDJ-SSW-JSC040), the China Postdoctoral Science Foundation (Grant No. 2021M693402). Jin Gao was also supported in part by the Youth Innovation Promotion Association, CAS.

{\small
\bibliographystyle{ieee_fullname}
\bibliography{egbib}

\begin{thebibliography}{10}
\providecommand{\url}[1]{#1}
\csname url@samestyle\endcsname
\providecommand{\newblock}{\relax}
\providecommand{\bibinfo}[2]{#2}
\providecommand{\BIBentrySTDinterwordspacing}{\spaceskip=0pt\relax}
\providecommand{\BIBentryALTinterwordstretchfactor}{4}
\providecommand{\BIBentryALTinterwordspacing}{\spaceskip=\fontdimen2\font plus
\BIBentryALTinterwordstretchfactor\fontdimen3\font minus
  \fontdimen4\font\relax}
\providecommand{\BIBforeignlanguage}[2]{{%
\expandafter\ifx\csname l@#1\endcsname\relax
\typeout{** WARNING: IEEEtran.bst: No hyphenation pattern has been}%
\typeout{** loaded for the language `#1'. Using the pattern for}%
\typeout{** the default language instead.}%
\else
\language=\csname l@#1\endcsname
\fi
#2}}
\providecommand{\BIBdecl}{\relax}
\BIBdecl

\bibitem{ren2015faster}
S.~Ren, K.~He, R.~Girshick, and J.~Sun, ``Faster r-cnn: Towards real-time
  object detection with region proposal networks,'' \emph{Advances in neural
  information processing systems}, vol.~28, pp. 91--99, 2015.

\bibitem{redmon2016you}
J.~Redmon, S.~Divvala, R.~Girshick, and A.~Farhadi, ``You only look once:
  Unified, real-time object detection,'' in \emph{Proceedings of the IEEE
  conference on computer vision and pattern recognition}, 2016, pp. 779--788.

\bibitem{redmon2017yolo9000}
J.~Redmon and A.~Farhadi, ``Yolo9000: better, faster, stronger,'' in
  \emph{Proceedings of the IEEE conference on computer vision and pattern
  recognition}, 2017, pp. 7263--7271.

\bibitem{redmon2018yolov3}
------, ``Yolov3: An incremental improvement,'' \emph{arXiv preprint
  arXiv:1804.02767}, 2018.

\bibitem{bochkovskiy2020yolov4}
A.~Bochkovskiy, C.-Y. Wang, and H.-Y.~M. Liao, ``Yolov4: Optimal speed and
  accuracy of object detection,'' \emph{arXiv preprint arXiv:2004.10934}, 2020.

\bibitem{Lin_2017_CVPR}
T.-Y. Lin, P.~Dollar, R.~Girshick, K.~He, B.~Hariharan, and S.~Belongie,
  ``Feature pyramid networks for object detection,'' in \emph{Proceedings of
  the IEEE Conference on Computer Vision and Pattern Recognition (CVPR)}, July
  2017.

\bibitem{Lin_2017_ICCV}
T.-Y. Lin, P.~Goyal, R.~Girshick, K.~He, and P.~Dollar, ``Focal loss for dense
  object detection,'' in \emph{Proceedings of the IEEE International Conference
  on Computer Vision (ICCV)}, Oct 2017.

\bibitem{tian2019fcos}
Z.~Tian, C.~Shen, H.~Chen, and T.~He, ``Fcos: Fully convolutional one-stage
  object detection,'' in \emph{Proceedings of the IEEE/CVF international
  conference on computer vision}, 2019, pp. 9627--9636.

\bibitem{Zhang_2020_CVPR}
S.~Zhang, C.~Chi, Y.~Yao, Z.~Lei, and S.~Z. Li, ``Bridging the gap between
  anchor-based and anchor-free detection via adaptive training sample
  selection,'' in \emph{Proceedings of the IEEE/CVF Conference on Computer
  Vision and Pattern Recognition (CVPR)}, June 2020.

\bibitem{paa-eccv2020}
K.~Kim and H.~S. Lee, ``Probabilistic anchor assignment with iou prediction for
  object detection,'' in \emph{ECCV}, 2020.

\bibitem{bansal2018zero}
A.~Bansal, K.~Sikka, G.~Sharma, R.~Chellappa, and A.~Divakaran, ``Zero-shot
  object detection,'' in \emph{Proceedings of the European Conference on
  Computer Vision (ECCV)}, 2018, pp. 384--400.

\bibitem{rahman2020improved}
S.~Rahman, S.~Khan, and N.~Barnes, ``Improved visual-semantic alignment for
  zero-shot object detection,'' in \emph{Proceedings of the AAAI Conference on
  Artificial Intelligence}, vol.~34, no.~07, 2020, pp. 11\,932--11\,939.

\bibitem{zareian2021open}
A.~Zareian, K.~D. Rosa, D.~H. Hu, and S.-F. Chang, ``Open-vocabulary object
  detection using captions,'' in \emph{Proceedings of the IEEE/CVF Conference
  on Computer Vision and Pattern Recognition}, 2021, pp. 14\,393--14\,402.

\bibitem{huang2020pixel}
Z.~Huang, Z.~Zeng, B.~Liu, D.~Fu, and J.~Fu, ``Pixel-bert: Aligning image
  pixels with text by deep multi-modal transformers,'' \emph{arXiv preprint
  arXiv:2004.00849}, 2020.

\bibitem{radford2021learning}
A.~Radford, J.~W. Kim, C.~Hallacy, A.~Ramesh, G.~Goh, S.~Agarwal, G.~Sastry,
  A.~Askell, P.~Mishkin, J.~Clark \emph{et~al.}, ``Learning transferable visual
  models from natural language supervision,'' \emph{arXiv preprint
  arXiv:2103.00020}, 2021.

\bibitem{gu2021zero}
X.~Gu, T.-Y. Lin, W.~Kuo, and Y.~Cui, ``Zero-shot detection via vision and
  language knowledge distillation,'' \emph{arXiv preprint arXiv:2104.13921},
  2021.

\bibitem{xie2021zsd}
J.~Xie and S.~Zheng, ``Zsd-yolo: Zero-shot yolo detection using vision-language
  knowledgedistillation,'' \emph{arXiv preprint arXiv:2109.12066}, 2021.

\bibitem{lin2014microsoft}
T.-Y. Lin, M.~Maire, S.~Belongie, J.~Hays, P.~Perona, D.~Ramanan,
  P.~Doll{\'a}r, and C.~L. Zitnick, ``Microsoft coco: Common objects in
  context,'' in \emph{European conference on computer vision}.\hskip 1em plus
  0.5em minus 0.4em\relax Springer, 2014, pp. 740--755.

\bibitem{chen2015microsoft}
X.~Chen, H.~Fang, T.-Y. Lin, R.~Vedantam, S.~Gupta, P.~Doll{\'a}r, and C.~L.
  Zitnick, ``Microsoft coco captions: Data collection and evaluation server,''
  \emph{arXiv preprint arXiv:1504.00325}, 2015.

\bibitem{demirel2018zero}
B.~Demirel, R.~G. Cinbis, and N.~Ikizler-Cinbis, ``Zero-shot object detection
  by hybrid region embedding,'' \emph{arXiv preprint arXiv:1805.06157}, 2018.

\bibitem{zhu2020don}
P.~Zhu, H.~Wang, and V.~Saligrama, ``Don't even look once: Synthesizing
  features for zero-shot detection,'' in \emph{Proceedings of the IEEE/CVF
  Conference on Computer Vision and Pattern Recognition}, 2020, pp.
  11\,693--11\,702.

\bibitem{rahman2019transductive}
S.~Rahman, S.~Khan, and N.~Barnes, ``Transductive learning for zero-shot object
  detection,'' in \emph{Proceedings of the IEEE/CVF International Conference on
  Computer Vision}, 2019, pp. 6082--6091.

\bibitem{farhadi2009describing}
A.~Farhadi, I.~Endres, D.~Hoiem, and D.~Forsyth, ``Describing objects by their
  attributes,'' in \emph{2009 IEEE conference on computer vision and pattern
  recognition}.\hskip 1em plus 0.5em minus 0.4em\relax IEEE, 2009, pp.
  1778--1785.

\bibitem{jayaraman2014zero}
D.~Jayaraman and K.~Grauman, ``Zero shot recognition with unreliable
  attributes,'' \emph{arXiv preprint arXiv:1409.4327}, 2014.

\bibitem{palatucci2009zero}
M.~M. Palatucci, D.~A. Pomerleau, G.~E. Hinton, and T.~Mitchell, ``Zero-shot
  learning with semantic output codes,'' 2009.

\bibitem{frome2013devise}
A.~Frome, G.~Corrado, J.~Shlens, S.~Bengio, J.~Dean, M.~Ranzato, and
  T.~Mikolov, ``Devise: A deep visual-semantic embedding model,'' 2013.

\bibitem{norouzi2013zero}
M.~Norouzi, T.~Mikolov, S.~Bengio, Y.~Singer, J.~Shlens, A.~Frome, G.~S.
  Corrado, and J.~Dean, ``Zero-shot learning by convex combination of semantic
  embeddings,'' \emph{arXiv preprint arXiv:1312.5650}, 2013.

\bibitem{he2020momentum}
K.~He, H.~Fan, Y.~Wu, S.~Xie, and R.~Girshick, ``Momentum contrast for
  unsupervised visual representation learning,'' in \emph{Proceedings of the
  IEEE/CVF Conference on Computer Vision and Pattern Recognition}, 2020, pp.
  9729--9738.

\bibitem{chen2020simple}
T.~Chen, S.~Kornblith, M.~Norouzi, and G.~Hinton, ``A simple framework for
  contrastive learning of visual representations,'' in \emph{International
  conference on machine learning}.\hskip 1em plus 0.5em minus 0.4em\relax PMLR,
  2020, pp. 1597--1607.

\bibitem{chen2021exploring}
X.~Chen and K.~He, ``Exploring simple siamese representation learning,'' in
  \emph{Proceedings of the IEEE/CVF Conference on Computer Vision and Pattern
  Recognition}, 2021, pp. 15\,750--15\,758.

\bibitem{he2016deep}
K.~He, X.~Zhang, S.~Ren, and J.~Sun, ``Deep residual learning for image
  recognition,'' in \emph{Proceedings of the IEEE conference on computer vision
  and pattern recognition}, 2016, pp. 770--778.

\bibitem{wang2018zero}
X.~Wang, Y.~Ye, and A.~Gupta, ``Zero-shot recognition via semantic embeddings
  and knowledge graphs,'' in \emph{Proceedings of the IEEE conference on
  computer vision and pattern recognition}, 2018, pp. 6857--6866.

\bibitem{zheng2021zero}
Y.~Zheng, J.~Wu, Y.~Qin, F.~Zhang, and L.~Cui, ``Zero-shot instance
  segmentation,'' in \emph{Proceedings of the IEEE/CVF Conference on Computer
  Vision and Pattern Recognition}, 2021, pp. 2593--2602.

\bibitem{li2019zero}
Z.~Li, L.~Yao, X.~Zhang, X.~Wang, S.~Kanhere, and H.~Zhang, ``Zero-shot object
  detection with textual descriptions,'' in \emph{Proceedings of the AAAI
  Conference on Artificial Intelligence}, vol.~33, no.~01, 2019, pp.
  8690--8697.

\bibitem{zheng2020background}
Y.~Zheng, R.~Huang, C.~Han, X.~Huang, and L.~Cui, ``Background learnable
  cascade for zero-shot object detection,'' in \emph{Proceedings of the Asian
  Conference on Computer Vision}, 2020.

\bibitem{zhao2020gtnet}
S.~Zhao, C.~Gao, Y.~Shao, L.~Li, C.~Yu, Z.~Ji, and N.~Sang, ``Gtnet: Generative
  transfer network for zero-shot object detection,'' in \emph{Proceedings of
  the AAAI Conference on Artificial Intelligence}, vol.~34, no.~07, 2020, pp.
  12\,967--12\,974.

\bibitem{sharma2018conceptual}
P.~Sharma, N.~Ding, S.~Goodman, and R.~Soricut, ``Conceptual captions: A
  cleaned, hypernymed, image alt-text dataset for automatic image captioning,''
  in \emph{Proceedings of the 56th Annual Meeting of the Association for
  Computational Linguistics (Volume 1: Long Papers)}, 2018, pp. 2556--2565.

\bibitem{saichev2009theory}
A.~I. Saichev, Y.~Malevergne, and D.~Sornette, \emph{Theory of Zipf's law and
  beyond}.\hskip 1em plus 0.5em minus 0.4em\relax Springer Science \& Business
  Media, 2009, vol. 632.

\bibitem{chen2019mmdetection}
K.~Chen, J.~Wang, J.~Pang, Y.~Cao, Y.~Xiong, X.~Li, S.~Sun, W.~Feng, Z.~Liu,
  J.~Xu \emph{et~al.}, ``Mmdetection: Open mmlab detection toolbox and
  benchmark,'' \emph{arXiv preprint arXiv:1906.07155}, 2019.

\bibitem{li2020generalized}
X.~Li, W.~Wang, L.~Wu, S.~Chen, X.~Hu, J.~Li, J.~Tang, and J.~Yang,
  ``Generalized focal loss: Learning qualified and distributed bounding boxes
  for dense object detection,'' \emph{Advances in Neural Information Processing
  Systems}, vol.~33, pp. 21\,002--21\,012, 2020.

\bibitem{li2021generalized}
X.~Li, W.~Wang, X.~Hu, J.~Li, J.~Tang, and J.~Yang, ``Generalized focal loss
  v2: Learning reliable localization quality estimation for dense object
  detection,'' in \emph{Proceedings of the IEEE/CVF Conference on Computer
  Vision and Pattern Recognition}, 2021, pp. 11\,632--11\,641.

\bibitem{Feng_2021_ICCV}
C.~Feng, Y.~Zhong, Y.~Gao, M.~R. Scott, and W.~Huang, ``Tood: Task-aligned
  one-stage object detection,'' in \emph{Proceedings of the IEEE/CVF
  International Conference on Computer Vision (ICCV)}, October 2021, pp.
  3510--3519.

\bibitem{zhou2021coop}
K.~Zhou, J.~Yang, C.~C. Loy, and Z.~Liu, ``Learning to prompt for
  vision-language models,'' \emph{arXiv preprint arXiv:2109.01134}, 2021.

\end{thebibliography}
}

\end{document}